\documentclass[10pt,journal,letterpaper]{IEEEtran}

\usepackage{cite}
\usepackage{amsmath,amssymb,amsfonts}
\usepackage{textcomp}
\usepackage{graphicx}
\usepackage[caption=false,font=normalsize,labelfont=sf,textfont=sf]{subfig}
\usepackage{array}
\usepackage{multirow}
\usepackage{algorithmic}
\usepackage{algorithm}
\usepackage{bm}
\usepackage{verbatim}
\usepackage{url}

\def\BibTeX{{\rm B\kern-.05em{\sc i\kern-.025em b}\kern-.08em
    T\kern-.1667em\lower.7ex\hbox{E}\kern-.125emX}}
    
\begin{document}

\title{Neural Window Decoding of Spatially Coupled LDPC Codes:\\
Exploiting Window Structure for Learning and Adaptation}
\author{Dae-Young~Yun,~\IEEEmembership{Member,~IEEE,}
Hee-Youl~Kwak,~\IEEEmembership{Member,~IEEE,}
Sang-Hyo~Kim,~\IEEEmembership{Member,~IEEE,}
Yongjune~Kim,~\IEEEmembership{Member,~IEEE,}
and~Jong-Seon~No,~\IEEEmembership{Life Fellow,~IEEE}%
\thanks{This work has been submitted to the IEEE for possible publication. Copyright may be transferred without notice, after which this version may no longer be accessible.}%
\thanks{A preliminary version of this work was presented at the 2024 KICS Winter Conference~\cite{Yun2024KICS}.}%
\thanks{Corresponding author: Hee-Youl Kwak (e-mail: ghy1228@gmail.com).}%
\thanks{Dae-Young Yun and Hee-Youl Kwak are with the Department of Electrical, Electronic and Computer Engineering, University of Ulsan, Ulsan 44610, South Korea (e-mail: dyyun95@gmail.com; ghy1228@gmail.com).}%
\thanks{Sang-Hyo Kim is with the Department of Electrical and Computer Engineering, Sungkyunkwan University, Suwon 16419, South Korea (e-mail: iamshkim@skku.edu).}%
\thanks{Yongjune Kim is with the Department of Electrical Engineering, Pohang University of Science and Technology (POSTECH), Pohang, Gyeongbuk 37673, South Korea (e-mail: yongjune@postech.ac.kr).}%
\thanks{Jong-Seon No is with the Department of Electrical and Computer Engineering and the Institute of New Media and Communications, Seoul National University, Seoul 08826, South Korea (e-mail: jsno@snu.ac.kr).}}

\markboth{Author preprint}%
{Yun \MakeLowercase{\textit{et al.}}: Neural Window Decoding of Spatially Coupled LDPC Codes}

\maketitle

\begin{abstract}
We propose a neural window decoder (NWD) for spatially coupled low-density parity-check (SC-LDPC) codes. The proposed methods adapt neural min-sum decoding to the structure and sequential operation of window decoding. First, target-oriented training restricts the loss to the variable nodes whose decisions are required at each window stage. Second, learned damping factors identify check-node updates that can reuse messages from the preceding iteration. Finally, adaptive weight switching selects weights trained for unreliable inherited information when a previous-window failure is detected. For a length-$20{,}000$ $(3,6)$-regular SC-LDPC code, the target-oriented NWD achieves performance comparable to sum-product window decoding while using min-sum updates. On the reduced target graph, scheduling based on regularized damping factors yields an estimated computational cost 48\% lower than that of full-graph weighted min-sum decoding, while maintaining comparable decoding performance. Adaptive weight switching improves decoding performance by 35\% at 3.0 dB relative to plain weights. These learning methods exploit the properties of window decoding to improve the reliability and computational efficiency of SC-LDPC convolutional codes.
\end{abstract}


\section{Introduction}

\IEEEPARstart{E}{fficient} receiver design for mobile and vehicular communications requires channel decoding that combines reliable error correction with manageable computation, memory requirements, and decoding latency. Iterative message-passing decoding of low-density parity-check (LDPC) codes offers a practical approach through local computations on sparse graphs, per-iteration complexity that scales linearly with block length for fixed node degrees, and opportunities for parallel implementation~\cite{Gallager1962,MacKay1999}. Its error-correction performance and implementation cost depend on both the message-update rules and the decoding effort allocated within a limited iteration budget.

Spatially coupled LDPC (SC-LDPC) codes combine strong iterative-decoding performance with a convolutional graph structure~\cite{Felstrom1999}. Through threshold saturation, spatial coupling can bring the belief-propagation (BP) threshold close to the maximum a posteriori threshold of the underlying uncoupled ensemble~\cite{Kudekar2011}. Sliding-window decoding exploits the local connectivity of these codes to process a limited portion of the chain at a time, reducing the latency and memory requirements of full-chain decoding~\cite{Iyengar2012,Kwak2022_SC}. This allows long SC-LDPC chains to be decoded using a smaller local graph, making efficient and reliable processing within each window a central design objective.

Existing scheduling methods reduce computation by allocating decoding effort non-uniformly across positions and iterations. These include dynamic schedules for LDPC convolutional codes~\cite{Weng2011}, reliability-based and pragmatic window schedules~\cite{Hassan2017}, and schedules optimized through protograph analysis and integer programming~\cite{Khittiwitchayakul2022}.
Window decoding is also susceptible to error propagation because erroneous decisions from one window can affect subsequent windows. Existing countermeasures adapt window operation~\cite{Klaiber2018,Griebel2024} or introduce check-node (CN) and variable-node (VN) doping~\cite{Zhu2023}.

To improve the performance of iterative decoders, neural decoding introduces learned parameters into an unrolled message-passing graph~\cite{Matsumine2024}. Early neural BP and min-sum (MS) decoders introduced trainable message weights while retaining the underlying decoding structure~\cite{Nachmani2016,Nachmani2018}. Weight sharing across iterations or groups of edges reduces the number of trainable parameters~\cite{Nachmani2017,Lian2019}. Neural decoding has been investigated for LDPC codes, including protograph-based constructions~\cite{Dai2021,Wang2024}, and generalized LDPC (GLDPC) codes~\cite{Kwak2022_GLDPC}. Learned weights have also been used to guide CN pruning in neural BP decoders with overcomplete parity-check matrices~\cite{Buchberger2021}.

Applying neural decoding to SC-LDPC window decoding requires accounting for three properties of window operation. Decisions are needed only for target VNs, the need for fresh message updates varies across positions and iterations, and errors in inherited log-likelihood ratios (LLRs) can propagate to subsequent windows. We therefore propose a neural window decoding (NWD) framework based on neural weighted min-sum (NWMS) decoding that incorporates these properties while preserving the code, window geometry, and iteration budget. The main contributions are as follows:

\begin{enumerate}
    \item \emph{Target-oriented training}: We align NWD training with the target decisions emitted at each window stage and reduce the computation graph to the message dependencies required for those decisions. For fixed weights and a fixed iteration budget, graph reduction preserves the target LLRs.

    \item \emph{Learned scheduling}: We train only the damping factors while keeping the NWD weights fixed to rank CN updates for message reuse. By varying the number of rested updates, the proposed scheme enables flexible selection across a wide range of complexity--performance tradeoffs. All schedules are derived from a single offline ranking, without online ranking or retraining the weights for each schedule.

    \item \emph{Adaptive weight switching}: We use an unsatisfied-check-node (UCN) indicator for the previous block to select plain weights or barrier weights trained on windows following decoding failures. This improves robustness to unreliable inherited LLRs without increasing the iteration budget or changing the message-update structure, with one additional weight set and a UCN-based selection rule.
\end{enumerate}

For the evaluated length-$20{,}000$ $(3,6)$-regular SC-LDPC code with window width $W=10$ and ten iterations per window, target-oriented NWD achieves performance comparable to sum-product window decoding while using MS updates. With 29 rested updates ($R=29$), the regularized damping schedule reduces the estimated computational cost in the reference window by approximately 48\% relative to full-graph weighted min-sum decoding while maintaining comparable performance at a frame-error rate of $10^{-1}$. In a separate evaluation, UCN-based adaptive weight switching reduces the average block-error rate by 35\% at 3.0 dB relative to plain weights.

The remainder of this paper is organized as follows. Section~\ref{Sec:Preliminaries} reviews SC-LDPC window decoding and NWMS decoding. Section~\ref{Sec:NWD} examines target-oriented training and its effects on decoding performance and computation. Sections~\ref{Sec:scheduling} and~\ref{Sec:ANWD} describe learned scheduling and adaptive weight switching, respectively. Section~\ref{Sec:Complexity} compares their computational and storage costs, and Section~\ref{Sec:Conclusion} concludes the paper.

\section{Preliminaries}\label{Sec:Preliminaries}

This section provides background on SC-LDPC code structure, the properties of window decoding, and prior methods that exploit these properties. We also review the NWMS message-passing equations.

\subsection{SC-LDPC Codes and Window Decoding}

\begin{figure}[t]
\centering
\subfloat[]{\includegraphics[width=7.5cm]{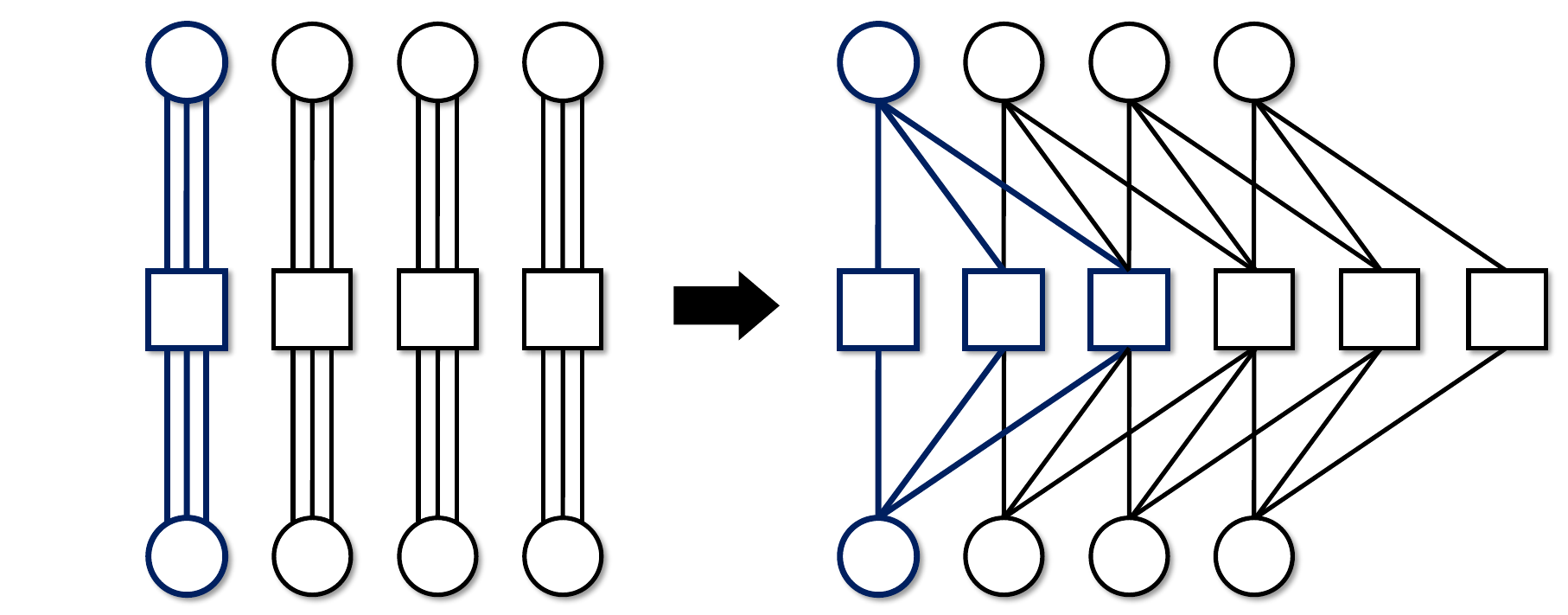}}\par
\subfloat[]{\includegraphics[width=8.0cm]{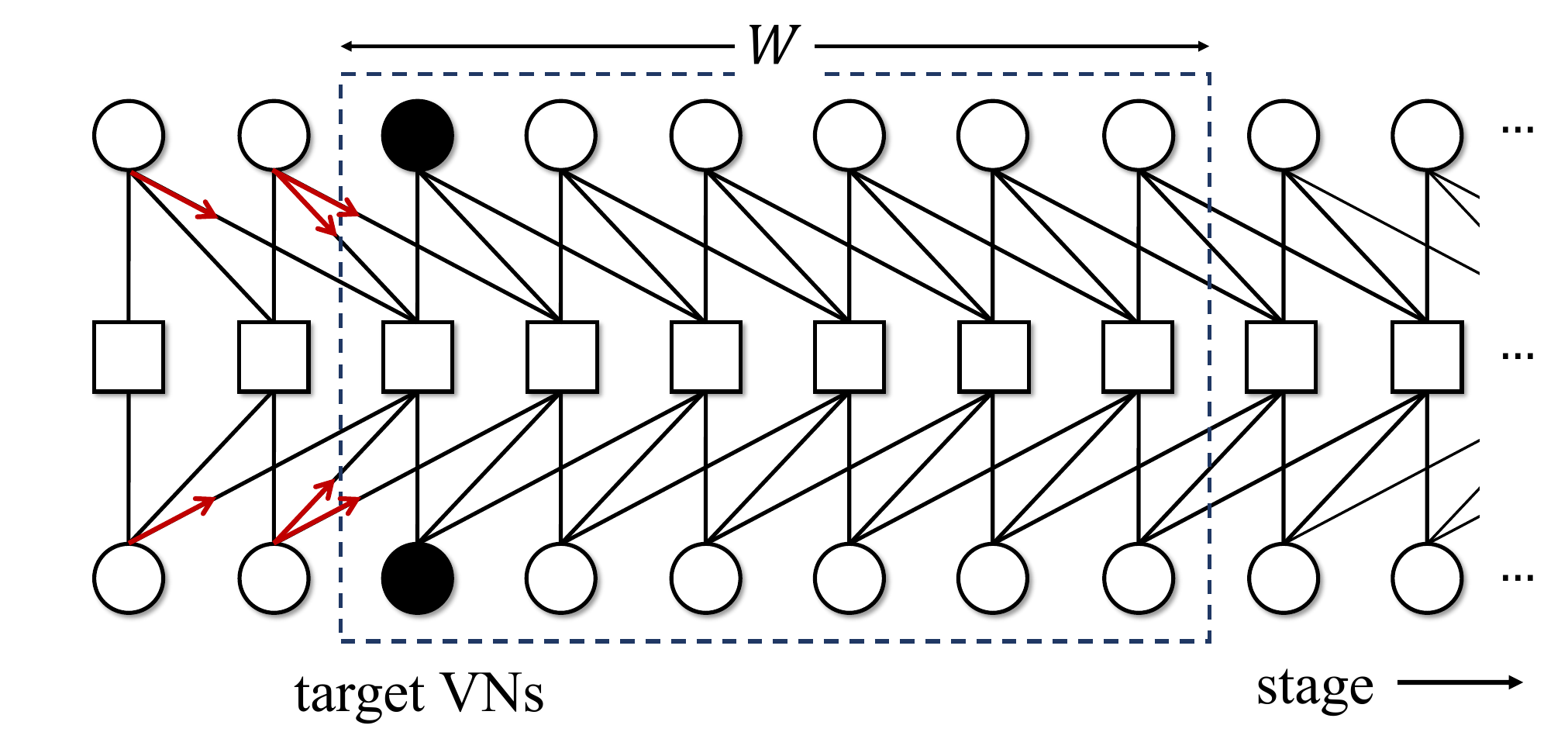}}
\caption{(a) Protograph construction of the terminated $(3,6)$-regular SC-LDPC code with $w=2$ and $L=4$ by edge spreading. (b) Window decoding with $W=6$ and one target position. The dotted box denotes the window at the third stage; the dark-filled VNs are the target VNs, and the arrows denote inherited LLRs from earlier decoded blocks.}
\label{Fig:WD}
\end{figure}

We consider terminated protograph-based SC-LDPC codes constructed by edge spreading~\cite{Mitchell2015}. Let $w$ denote the coupling width and $L$ the chain length. The base matrix has VNs at $L$ positions and CNs at $L+w$ positions, with $N$ protograph VNs and $M$ protograph CNs per position. Fig.~\ref{Fig:WD}(a) shows the $(3,6)$-regular, $w=2$ protograph used throughout the experiments. Lifting by $z$ gives $n=Nz$ VNs and $m=Mz$ CNs per position, code length $Ln$, and design rate
\begin{equation*}
  \mathcal{R}=1-\frac{(L+w)m}{Ln}=\bigg(1-\frac{M}{N}\bigg)-\frac{M}{N}\frac{w}{L}.
\end{equation*}

Sliding-window decoding operates on $W$ positions, decides the target nodes in the first position, and then shifts one position to the right~\cite{Iyengar2012}, as illustrated in Fig.~\ref{Fig:WD}(b). From the second stage onward, previously decoded VNs connected to the window supply inherited LLRs that remain fixed throughout its iterations.
Previously decoded VNs emit their saved target-output LLRs as fixed VN-to-CN messages, without adding new CN-to-VN feedback. We reset internal CN-to-VN messages to zero at each window shift. At the terminated right boundary, we retain only the nodes and edges present in the full parity-check matrix.

The simulations use binary phase-shift keying over an additive white Gaussian noise (AWGN) channel. The block-error rate (BLER) is the decoding failure rate of the target block at each window position, whereas the frame-error rate (FER) is the decoding failure rate of the entire length-$Ln$ chain.

\subsection{Non-Uniform Window Decoding Schedules}

In SC-LDPC window decoding, only the VNs at the first window position require decisions; the remaining nodes support these target decisions through message passing. The contribution of an update to the target decisions therefore varies with its position and iteration. Non-uniform schedules use this variation to allocate decoding effort selectively.
Conventional flooding schedules update all VNs and CNs in every iteration. Non-uniform schedules can vary the update order and the number of updates assigned to each node, providing a tradeoff between decoding performance and computation~\cite{Sharon2009,Elidan2006,Casado2010}.

For LDPC convolutional codes with rational parity-check matrices, dynamic scheduling on an equivalent Tanner graph with finite memory is combined with signal perturbation and error cancellation to improve convergence~\cite{Weng2011}. For SC-LDPC window decoding, non-uniform schedules select active positions according to the improvement in their soft bit-error rate (BER) estimates~\cite{Hassan2017}. Computing these estimates introduces online work. A pragmatic variant avoids this cost through a predetermined periodic pattern that progressively concentrates updates toward the target position.

Integer-programming schedules optimize VN updates using protograph extrinsic-information-transfer (PEXIT) analysis for a selected protograph, window width, and iteration budget~\cite{Khittiwitchayakul2022}. They allow the desired update reduction or threshold loss to be specified, but require analysis and optimization for the chosen configuration. Thus, existing schedules offer different tradeoffs between online estimation, prescribed update patterns, and model-based schedule construction.

\subsection{Error Propagation in Window Decoding}

In window decoding, the output LLRs of previously decoded target VNs serve as fixed inputs to subsequent windows. When the previous block is decoded correctly, the inherited messages support decoding of the current block. A decoding failure can instead pass erroneous messages to the current window, increasing the likelihood of another block error. Repeated failures can develop into a long burst, known as error propagation. This sequential dependence has been studied in window decoding of braided convolutional codes~\cite{Zhu2018,Zhu2020} and SC-LDPC codes~\cite{Zhu2023}.

Several approaches have been investigated to address this dependence. Stall-aware decoders can increase the iteration count locally or revisit earlier window positions through adaptive shifts~\cite{Klaiber2018}. Related adaptive window decoders use partial-syndrome information to control processor activation and the active window width~\cite{Griebel2024}. Although these adaptations can reduce average computation or energy consumption, they require data-dependent control of decoding effort and window operation.

CN doping introduces additional reduced-degree checks at selected positions, whereas VN doping fixes selected bits to known values. Both help the decoder recover from error propagation, but modify the code constraints and incur a small rate cost; adaptive doping additionally uses feedback to determine when doping is needed~\cite{Zhu2023}. These tradeoffs motivate examining error-propagation mitigation that preserves the code and message-passing structure within a fixed window and iteration budget.

\subsection{Neural Weighted Min-Sum Decoding}

Neural weighted min-sum (NWMS) decoding learns weights that scale CN-to-VN messages in an unrolled min-sum decoder~\cite{Nachmani2018}.
The channel LLR of VN $v$ is
\begin{equation}
m_v^{\rm ch}=\log\frac{\Pr(x_v=0\mid y_v)}{\Pr(x_v=1\mid y_v)},
\label{Eq:channel_llr}
\end{equation}
where $x_v$ is the code bit and $y_v$ is its channel observation. Positive LLRs favor bit 0, and negative LLRs favor bit 1.
Let $m^{(\ell)}_{v \rightarrow c}$ ($m^{(\ell)}_{c \rightarrow v}$) denote the message passed from VN $v$ to CN $c$ (and vice versa) during the $\ell$-th iteration. Initially, $m^{(0)}_{v\rightarrow c}=m_v^{\rm ch}$ and $m^{(0)}_{c\rightarrow v}=0$. For iteration $\ell=1,2,...,\overline{\ell}$, NWMS updates the messages as follows:
\begin{align}
        m^{(\ell)}_{v\rightarrow c} &= m_{v}^{\rm ch} + \sum_{c'\in \mathcal{N}(v) \setminus c}  m^{(\ell - 1)}_{c'\rightarrow v} \label{Eq:NWMS_VN}\\
    \begin{split}
    m^{(\ell)}_{c\rightarrow v} &= {w}^{(\ell)}_{c\rightarrow v}
    \prod_{v'\in \mathcal{N}(c) \setminus v} {\rm sgn}\left(m^{(\ell)}_{v'\rightarrow c}\right)\\
    &\quad{}\times \min_{v'\in \mathcal{N}(c) \setminus v}|m^{(\ell)}_{v'\rightarrow c}|,
    \end{split}
    \label{Eq:NWMS_CN}
\end{align}
where $\mathcal{N}(v)$ and $\mathcal{N}(c)$ denote the neighbors of VN $v$ and CN $c$, respectively. We refer to ${w}^{(\ell)}_{c\rightarrow v}$ as CN weights as they scale CN messages. Fixing these weights to a common value yields the conventional weighted min-sum (WMS) decoder. At the last iteration, the output LLR $m_v^{\rm o}$ of VN $v$ is calculated as
\begin{align}
m_{v}^{\rm o} = m_{v}^{\rm ch} + \sum_{c'\in \mathcal{N}(v)} m^{ (\overline{\ell})}_{c'\rightarrow v}. %
\end{align}
In our implementation, the output LLRs are clipped to $[-20,20]$.

To reduce the number of trainable parameters and improve training efficiency, we use protograph weight sharing~\cite{Dai2021} and CN weight sharing~\cite{Dai2021}. This gives one weight $w_C^{(\ell)}$ per protograph CN $C$ and iteration $\ell$. We denote the complete set of trainable weights by $\mathcal{W}=\{w_C^{(\ell)}\}_{C,\ell}$, which contains $WM\overline{\ell}$ parameters for the full window graph. The weights are indexed relative to the window's left edge and reused after each shift.

\section{Neural Window Decoding for SC-LDPC Codes}\label{Sec:NWD}

At each window stage, decisions are required only for the target VNs. This provides a common focus for both learning and computation: the training objective can be restricted to the target outputs, and message dependencies can be used to identify the computations required to produce them. This section examines the resulting decoding performance and computational savings.

\begin{figure*}[!ht]
\centering
\includegraphics[width=17.0cm]{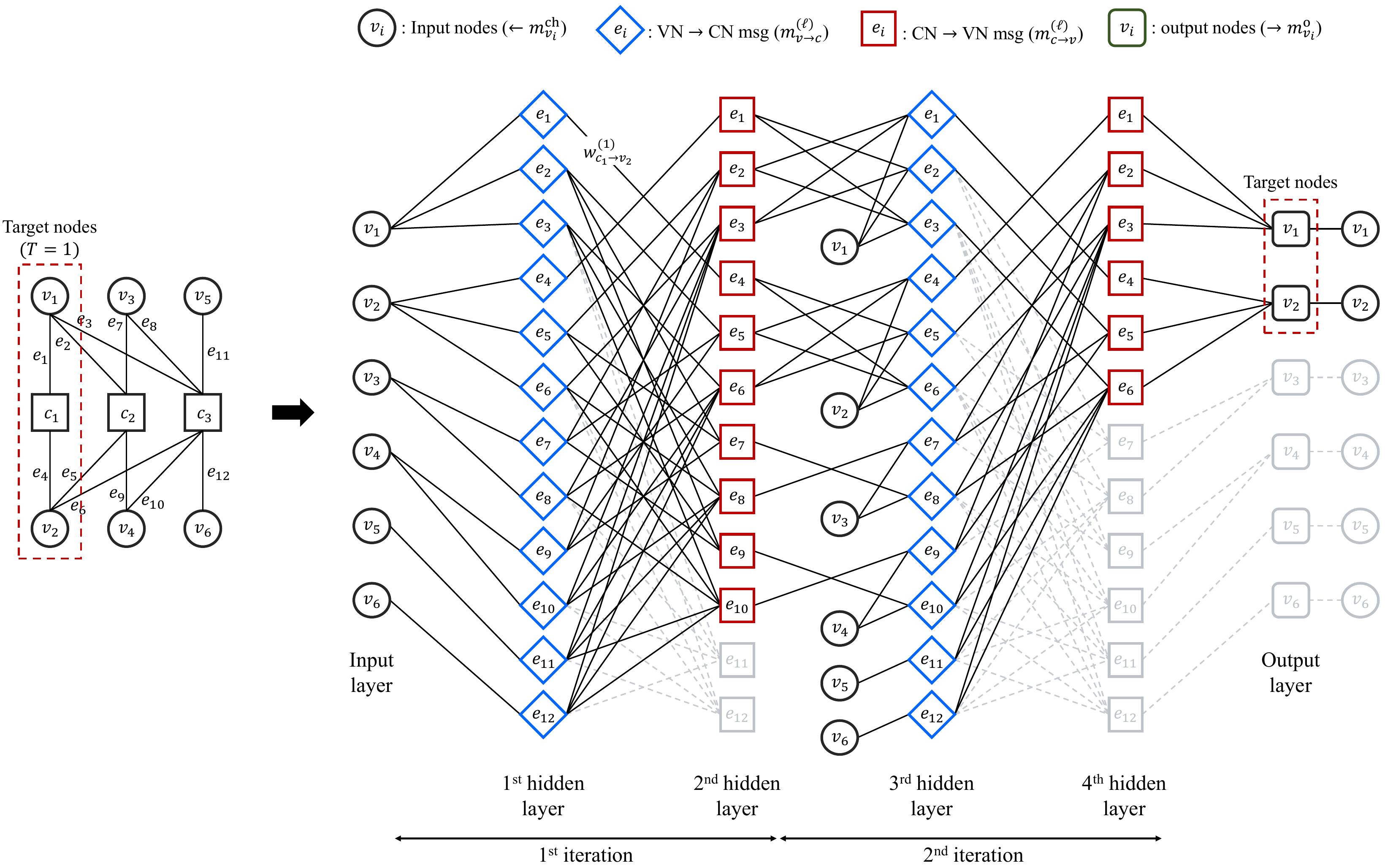}
\caption{The terminated first window with $W=3$ and $z=1$, unrolled for $\overline{\ell}=2$ iterations. Solid connections are retained for the target outputs; light-gray nodes and dashed connections can be omitted. The labels $e_i$ identify Tanner-graph edges and their corresponding directed messages. Repeated input circles denote the same channel LLRs used to compute the VN updates and output LLRs.}
\label{Fig:NWD}
\end{figure*}

\subsection{Target-Oriented Training}

Fig.~\ref{Fig:NWD} shows the full computation graph of the terminated first window with $W=3$, $z=1$, and $\overline{\ell}=2$ iterations. The graph has $Wn$ distinct channel inputs and $Wn$ outputs, and its hidden-layer nodes represent Tanner-graph edges. Odd and even hidden layers perform the VN and CN updates in (\ref{Eq:NWMS_VN}) and (\ref{Eq:NWMS_CN}), respectively. Channel inputs are repeated in the drawing to show their contributions to both VN-update layers and the output LLRs.
The same local decoder is reused as the window advances along the chain. We therefore train the weights on a single independent window and apply them at subsequent stages. This permits training on a small window graph while decoding a much longer SC-LDPC chain.

To examine the effect of the training objective, we refer to the baseline trained with a block-error loss over all $Wn$ outputs of the full window graph as the neural block decoder (NBD). Both NBD and the proposed NWD operate in a sliding-window manner, while NWD restricts the training loss to the target outputs and uses the reduced graph described below. For the all-zero training codeword, the block-error loss of NBD is~\cite{Xiao2021}
\begin{align}
L_{\rm A}
&= \frac{1}{2}\left[ 1 - {\rm sgn}\left( \mathop{\rm min}\limits_{1 \le v \le Wn} m_v^{\rm o} \right) \right].
\end{align}

Even when all target decisions are correct, errors in non-target outputs can keep the all-output loss nonzero. Since non-target outputs will be decided after later window shifts, target-oriented training includes only the first $Tn$ outputs, where $T$ is the number of target positions in the loss. We use $T=1$ to match the block decoded at each stage:
\begin{align}
L_{\rm T}
&= \frac{1}{2}\left[ 1 - {\rm sgn}\left( \mathop{\rm min}\limits_{1 \le v \le Tn} m_v^{\rm o} \right) \right].
\label{Eq:FER_loss}
\end{align}

\begin{figure}[t]
\centering
\includegraphics[width=0.85\columnwidth]{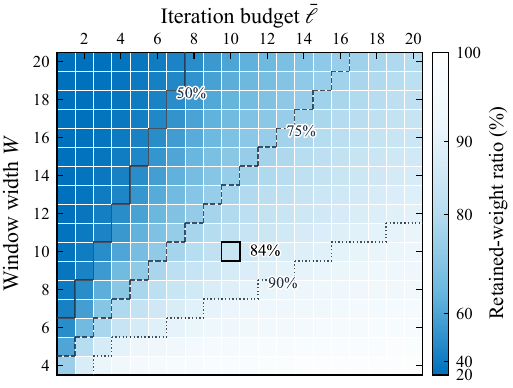}
\vspace{0.3cm}
\caption{Retained-weight ratio of NWD relative to NBD for the $(3,6)$-regular, $w=2$, $T=1$ graph. Each cell shows one combination of iteration budget $\overline{\ell}$ and window width $W$; darker blue denotes a smaller retained fraction and hence greater reduction. Step boundaries indicate the 50\%, 75\%, and 90\% retained-weight thresholds. The outlined cell marks the main setting $W=10$, $\overline{\ell}=10$.}
\label{Fig:Weight_ratio}
\end{figure}

\begin{figure*}[t]
\centering
\makebox[\textwidth][l]{%
  \hspace*{\dimexpr0.03\textwidth-3mm\relax}%
  \includegraphics[width=0.94\textwidth]{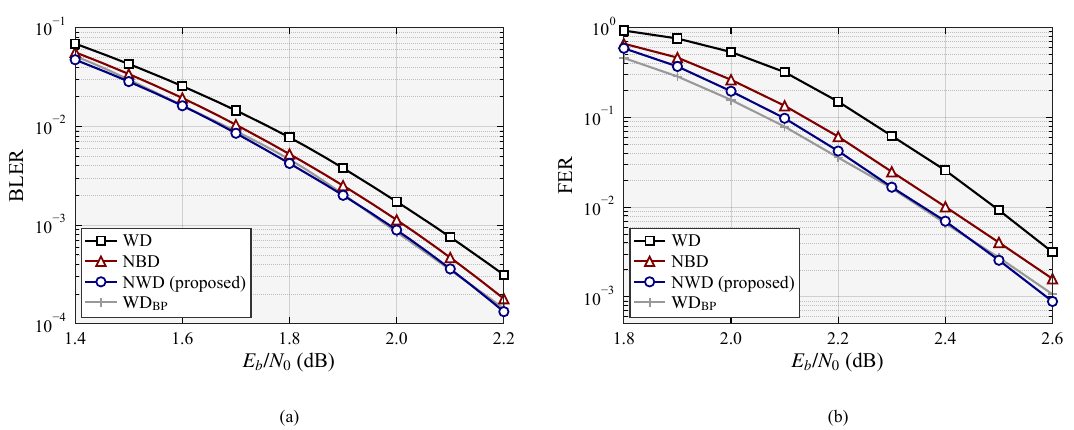}%
}
\caption{Error-rate performance for $w=2$, $z=100$, $W=10$, $T=1$, and $\overline{\ell}=10$: (a) single-window BLER and (b) full-chain FER for $L=100$ ($Ln=20{,}000$). WD, NBD, and NWD use MS-based updates; only WD\textsubscript{BP} uses sum-product BP updates. WD uses one fixed shared weight, NBD is trained on all window outputs, and NWD is trained on the target outputs and retains the target subgraph.}
\label{Fig:T_performance}
\end{figure*}

When only the target outputs are required, computations unrelated to them can be omitted regardless of the training objective. Starting from the final target LLRs, we recursively retain every input to each required VN or CN update until reaching the channel and inherited LLRs. The omitted connections are dashed in Fig.~\ref{Fig:NWD}. For fixed weights and a fixed iteration budget, this reduction leaves the target LLRs unchanged; non-target nodes that support the target decisions remain in the graph. For $T=1$ in Fig.~\ref{Fig:NWD}, the graph retains 48 of 60 nodes and 100 of 146 connections, corresponding to reductions of $20\%$ and approximately $32\%$, respectively.

Fig.~\ref{Fig:Weight_ratio} shows the retained-weight ratio jointly over $W$ and $\overline{\ell}$. A larger iteration budget expands the target subgraph farther into the window, whereas a wider window places more weights outside this subgraph. At the main $W=10$, $\overline{\ell}=10$ setting, NWD retains 84\% of the NBD weights; at $W=18$ and $\overline{\ell}=4$, it retains one third. Thus, low-latency window decoding with wider windows and fewer iterations~\cite{Kwak2022_SC,Huang2015} benefits from a greater reduction in the number of trainable weights.

\subsection{Training and Weight Selection}

Equal raw sample counts across signal-to-noise ratios (SNRs) can bias learning toward low SNR, where decoding failures and nonzero block losses occur more often. We therefore use active sample collection~\cite{Beery2020}: each mini-batch contains equal numbers of received words that the current decoder fails to decode at each training SNR. Samples are newly collected for every mini-batch.
We select the final weights using normalized validation error (NVE)~\cite{Gruber2017}. At each training SNR, the mean validation loss is divided by its fixed epoch-zero value, and the normalized losses are averaged across the SNRs. Validation uses ordinary, unconditioned AWGN realizations regenerated at each check. The parameters yielding the lowest NVE are selected as optimal.
For the block loss, we use a straight-through estimator~\cite{Bengio2013}. The forward pass evaluates the hard sign, while the backward pass replaces only its derivative with that of $\tanh(x/2)$. With ${\rm sgn}(0)=0$, the forward loss equals $1/2$ at a zero decision margin.

\subsection{Experimental Setup and Decoding Performance}

We use a terminated $(3,6)$-regular SC-LDPC code with chain length $L=100$, lifting factor $z=100$, and total code length $LNz=20{,}000$. For this code with coupling width $w=2$, the window decoder operates from left to right with $W=10$, $T=1$, and $\overline{\ell}=10$.
The NWD weights are trained on an independent window at the terminated left boundary. The training SNRs are $\mathcal{S}=\{1.2,1.4,1.6,1.8,2.0\}$ dB.
For both NBD and NWD, CN weights start at 0.75~\cite{Kwak2024,Kwak2025} and are optimized by Adam~\cite{Kingma2014} with a constant learning rate $10^{-3}$ and projection onto $[0,2]$. The conventional WMS decoder uses a fixed weight of 0.75~\cite{Kwak2024,Kwak2025}. NWD is trained for 1000 epochs, each containing ten mini-batches with 20 failed samples per SNR. NVE is evaluated after each epoch on 50,000 samples, 10,000 per SNR.
Single-window BLER simulations stop after 500 target-block errors or $10^7$ trials. Full-chain FER simulations collect 500 frame errors.

Fig.~\ref{Fig:T_performance}(a) compares NWD and NBD with the conventional WMS window decoder (WD) and the sum-product BP window decoder WD\textsubscript{BP}. NWD improves on both WD and NBD. Near BLER $=3\times10^{-4}$, its gain over WD is approximately 0.1 dB, and its performance is close to WD\textsubscript{BP}.
The full-chain result in Fig.~\ref{Fig:T_performance}(b) preserves this ordering. At FER~$=10^{-2}$, NWD gains approximately 0.13 dB over WD. At 2.5 dB, the FER decreases from $9.25\times10^{-3}$ for WD to $2.56\times10^{-3}$ for NWD, while WD\textsubscript{BP} gives $2.74\times10^{-3}$. The single-window and full-chain comparisons support matching the training loss to the decisions emitted at each window stage.

\begin{figure*}[t]
\centering
\includegraphics[width=\textwidth]{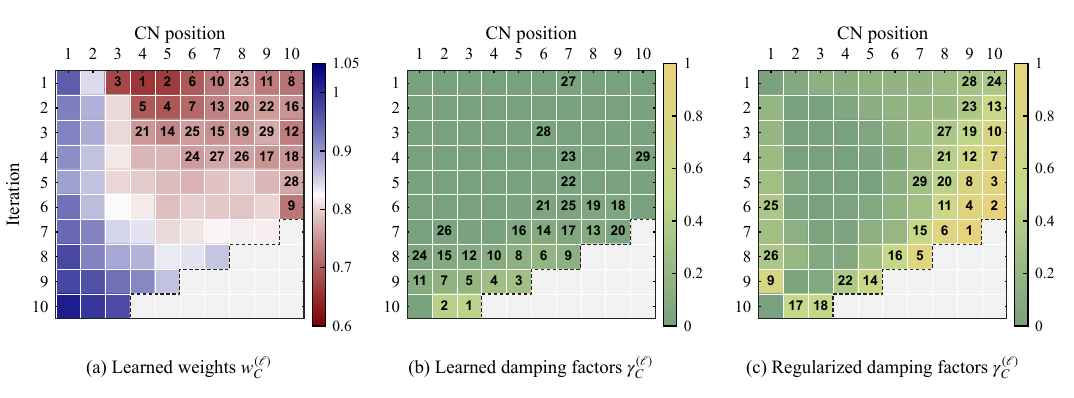}
\par\vspace{2.0pt}
\caption{CN-update ranking metrics for $W=10$, $T=1$, and $\overline{\ell}=10$: (a) NWD weights, (b) damping factors learned with $\lambda=0$, and (c) regularized damping factors with $\lambda=1$. Panels (b) and (c) train only damping factors while fixing the weights in (a). Numbers indicate the first 29 updates in each rest order. Gray cells lie outside the target subgraph.}
\label{Fig:damp}
\end{figure*}

\section{Learned Non-Uniform Scheduling}\label{Sec:scheduling}

Even after computations unrelated to the target outputs have been removed, the retained CN updates differ in the importance of computing fresh messages. We exploit these differences to reduce computation further within the fixed iteration budget. The proposed learned non-uniform schedule uses an offline ranking to select updates that reuse their preceding-iteration messages, concentrating fresh computation on updates assigned higher importance. Here, one CN update refers to the outgoing-message update of one protograph CN $C$ at iteration $\ell$, before lifting by $z$.
Unlike soft-BER-based schedule selection~\cite{Hassan2017}, our method requires no online reliability estimation to select updates. Instead of prescribing a periodic pattern or imposing PEXIT constraints~\cite{Hassan2017,Khittiwitchayakul2022}, we learn the rest priority from the target-block loss with the NWD weights fixed. For a given window geometry and iteration budget, a single offline ranking provides schedules with different rested-update counts without retraining the weights or performing online ranking.

\subsection{Damping Factors and Update Ranking}

Weight-based neural pruning uses learned CN weights to select iteration-dependent subsets of checks from an overcomplete parity-check matrix~\cite{Buchberger2021}. Our scheduling problem concerns whether to compute a fresh message or reuse the preceding-iteration message for each retained CN update. For comparison, we consider a weight-based rest schedule that selects the updates with the smallest trained NWD weights $w_C^{(\ell)}$ first. These weights scale CN messages, however, and do not directly indicate whether a fresh update is needed. A small weight can attenuate an unreliable but useful message, while a large weight can scale a message that changes little between iterations.

Damping combines fresh and preceding-iteration messages~\cite{Fossorier2003}. Learned damping has been used in weighted BP~\cite{Lian2019}. We propose a scheduling method that learns only the damping factors while keeping the trained NWD weights fixed. These factors are then used to rank the importance of fresh CN updates, so that less important updates can be rested during decoding by reusing their preceding-iteration messages. Let $\widehat{m}_{c\rightarrow v}^{(\ell)}$ be the fresh CN message from (\ref{Eq:NWMS_CN}). We introduce trainable damping factors into an auxiliary decoder:
\begin{align}
m_{c\rightarrow v}^{(\ell)}
={}&\gamma_C^{(\ell)}m_{c\rightarrow v}^{(\ell-1)}
 +(1-\gamma_C^{(\ell)})\widehat{m}_{c\rightarrow v}^{(\ell)},
\label{Eq:damp}
\end{align}
where $0\leq\gamma_C^{(\ell)}\leq1$ is shared across the outgoing messages of protograph CN $C$. The trained NWD weights remain fixed, and only the damping factors are optimized. A larger factor gives more weight to the previous message; $\gamma_C^{(\ell)}=1$ replaces the fresh update entirely. We therefore rank updates for resting in descending order of $\gamma_C^{(\ell)}$.

Without regularization, training only the damping factors using the decoding loss leaves most factors close to zero, making it difficult to distinguish the relative importance of CN updates. We therefore add a regularization term that encourages larger damping factors and hence greater reuse of preceding-iteration messages:
\begin{align}
\mathcal{L}_{\lambda}
={}&\mathbb{E}_{\rm batch}[L_{\rm T}]
 +\frac{\lambda}{WM\overline{\ell}}
   \sum_{\ell=1}^{\overline{\ell}}\sum_{C=1}^{WM}
   \bigl(1-\gamma_C^{(\ell)}\bigr).
\label{Eq:damp_loss}
\end{align}
Here, $\mathbb{E}_{\rm batch}$ denotes the average over the mini-batch. The unregularized and regularized damping methods use regularization factors $\lambda=0$ and $\lambda=1$, respectively.

Both damping fits start from $\gamma=0$ with fresh optimizer state and use 1000 Adam updates with learning rate $10^{-3}$ and zero weight decay. The factors are projected onto $[0,1]$ after each update. Each mini-batch contains 20 samples that the current damped decoder fails to decode at each SNR in $\{1.6,1.7,1.8,1.9,2.0\}$ dB.
The unregularized fit selects the checkpoint with the lowest pooled validation BLER. The regularized fit selects the checkpoint with the largest mean damping over the 84 retained updates, subject to a pooled validation BLER no greater than 1.2 times that of the fixed-weight decoder without damping.

Fig.~\ref{Fig:damp} compares the three ranking metrics. Weight-based scheduling primarily selects small weights in early iterations. Without regularization, most damping factors cluster near zero, making differences in update importance difficult to distinguish. Regularization spreads the factors over a wider range and produces distinctly large values close to one, making the rest priorities easier to distinguish. The resulting ranking identifies updates in the lower-right part of the retained graph, corresponding to larger CN positions and later iterations, as less important. This pattern is consistent with the triangular update pattern of the pragmatic schedule in~\cite{Hassan2017}, which progressively concentrates updates toward the target position. For $W=\overline{\ell}=10$, that schedule rests 29 of the 84 retained updates. To match this rest count, Fig.~\ref{Fig:damp} labels the 29 updates ranked least important by each metric in their rest order.

\subsection{Non-Uniform Schedules and Decoding Performance}

Using the rest orders in Fig.~\ref{Fig:damp}, we construct nested static schedules by resting the first $R$ updates, for $R=0,\ldots,29$. Here, $R=0$ denotes NWD without additional scheduling. Exact ties are resolved by later iteration and then larger CN position.
During decoding, the binary mask $a_C^{(\ell)}$ equals 1 for active updates and 0 for rested updates. Active updates follow (\ref{Eq:NWMS_CN}) with the original NWD weights, whereas rested updates reuse the preceding CN-to-VN messages. Each schedule is stored as its binary mask and active weights; the learned damping factors are discarded. VN-to-CN messages needed solely by rested updates can also be omitted, while reused CN messages remain available to neighboring active updates and target-output sums.

\begin{figure}[t]
\centering
\includegraphics[width=\columnwidth]{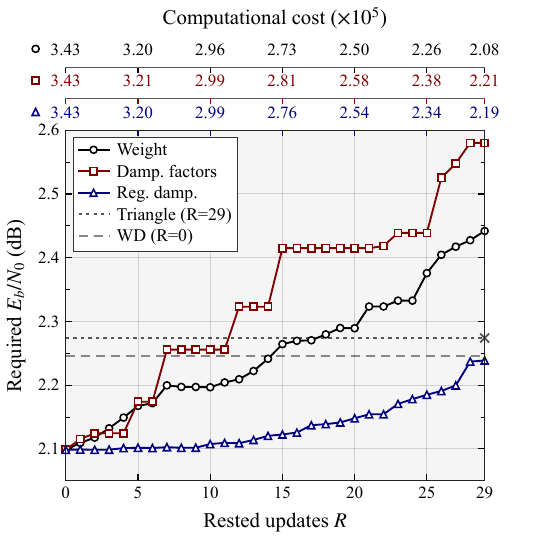}
\caption{Required $E_b/N_0$ at FER $=10^{-1}$ versus rested updates $R$ for three scheduling methods. The upper rows show computational costs; dotted and dashed lines denote Triangle ($R=29$) and WD ($R=0$), respectively.}
\label{Fig:scheduling}
\end{figure}

Fig.~\ref{Fig:scheduling} compares the estimated required SNR at FER $=10^{-1}$ for the three ranking methods with $L=100$, $z=100$, $W=10$, $T=1$, and $\overline{\ell}=10$. All methods use the same NWD weights, which remain fixed during damping training. As a reference, we include the pragmatic triangular schedule of~\cite{Hassan2017} with the same decoder and weights. Its active CN positions satisfy $C\leq W-\ell+1$, leaving 55 active updates at $R=29$. The dotted line represents this single schedule as a horizontal reference across $R$. For each schedule, every measured SNR point collects 500 frame errors, and the required SNR is obtained by interpolation in $\log_{10}(\mathrm{FER})$ between adjacent points spaced by 0.05 dB.

Regularized damping gives the lowest estimated required SNR among the three ranking methods for $R=1,\ldots,29$. At $R=8$, the interpolated change from NWD without scheduling is less than 0.01 dB, while 24 of the original 100 updates are omitted, including the 16 removed by graph reduction. At $R=25$, the required SNR is approximately 2.18 dB, compared with 2.38 dB for weight-based scheduling and 2.44 dB for unregularized damping. The regularized schedule incurs about 0.09 dB relative to the 2.10-dB reference at $R=0$ while omitting 41 updates in total.

At the matched rest count $R=29$, regularized damping requires an estimated 2.239 dB, compared with 2.274 dB for the pragmatic triangular reference. The full-graph WD reference, shown by the dashed line in Fig.~\ref{Fig:scheduling}, requires approximately 2.246 dB. Thus, the regularized schedule maintains comparable performance at this operating point while reducing the estimated computational cost by approximately 48\% relative to full-graph WD. The upper rows report the equivalent computational costs of the three ranking methods under the model in Section~\ref{Sec:Complexity}. These costs can differ at the same rest count because the positions of rested updates determine which message computations can be omitted.
The rest count therefore provides a configurable design parameter for selecting a complexity--performance operating point from a single learned ranking, without changing the window geometry or iteration budget.

The plateaus for unregularized damping over $R=7$--11 and $R=15$--21 arise because resting some late updates makes earlier computations unnecessary for the target outputs. Resting these already-unused updates leaves the effective graph unchanged. Thus, the nominal operation count can decrease even though an implementation that removes unused computations would perform the same operations throughout each plateau.

More broadly, the proposed damping-based ranking procedure does not rely on spatial coupling or a window boundary and can therefore be applied to other unrolled neural BP decoders. It is particularly well suited to decoding settings with structural asymmetry in update importance, where less important updates can be selectively rested to reduce computation. Its performance--complexity tradeoffs on those graphs remain to be evaluated.

\begin{figure*}[!t]
\centering
\includegraphics[width=\textwidth]{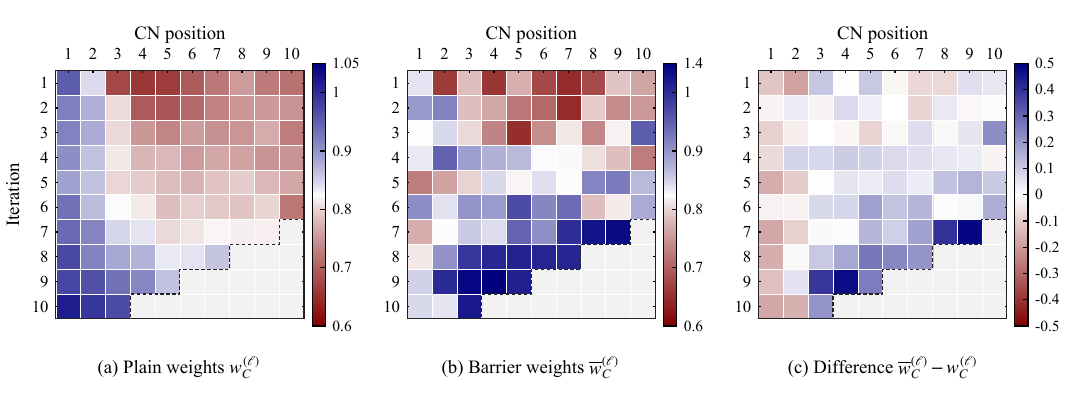}
\par\vspace{2.0pt}
\caption{Weight sets for adaptive weight switching: (a) plain weights $w_C^{(\ell)}$, (b) barrier weights $\overline{w}_C^{(\ell)}$, and (c) their difference $\overline{w}_C^{(\ell)}-w_C^{(\ell)}$.}
\label{Fig:conditioned_weight}
\end{figure*}

\section{Adaptive Weight Switching}\label{Sec:ANWD}

Reliable window decoding must account for information inherited from preceding windows. The plain NWD weights are trained on an independent window at the terminated left boundary, without inherited LLRs from previous decoding stages. They are therefore not explicitly optimized for the erroneous inherited information encountered after a decoding failure. To address this mismatch, we introduce adaptive switching between the plain weights and barrier weights trained on windows following previous-block errors. This adaptation aims to limit subsequent block errors while preserving the window geometry, iteration budget, and message-update structure.
In contrast to recovery through local iteration increases or window revisits~\cite{Klaiber2018} and adaptation of the active window width~\cite{Griebel2024}, weight switching preserves the window geometry, shift rule, iteration budget, and message-update structure. Unlike doping~\cite{Zhu2023}, it requires no changes to the code constraints or code rate. No extra message-passing updates are introduced; the additional online operations are a UCN test and weight-set selection, with one extra weight set stored.

Let $E_s$ denote the target-block error event at stage $s$. Following~\cite{Zhu2023}, we use $q_1=\Pr(E_s\mid E_{s-1})$ to measure the probability of a block error after an error in the previous block. The adaptive decoder aims to reduce this probability by changing the weights when a previous-block error is detected.

\subsection{Training the Barrier Weights}

We refer to the NWD weights from Section~\ref{Sec:NWD} as the plain weights $w_C^{(\ell)}$. To obtain the barrier weights $\overline{w}_C^{(\ell)}$, we collect windows following a previous-block error and retain both their channel LLRs and the inherited LLRs produced by the preceding decoding stages. Samples are selected by the previous-block outcome, regardless of whether the current block is decoded correctly. This training uses failure-conditioned samples, motivated by~\cite{Kwak2024,Kwak2025}.
The barrier weights are trained for 1000 epochs on 5000 cases collected at 2.0 dB. They start at 0.8 and use Adam with learning rate $10^{-3}$, batch size 20, and projection onto $[0,2]$, with the same block-loss surrogate as in Section~\ref{Sec:NWD}. The fixed sample pool is reused across epochs, and the checkpoint with the smallest empirical block loss on that pool is retained.

Fig.~\ref{Fig:conditioned_weight} compares the two weight sets and their difference. Relative to the plain set, the barrier set reduces the weights at CN position 1 across all iterations and increases several mid-to-rear weights in later iterations. This redistribution is consistent with attenuating unreliable inherited information while emphasizing messages developed within the current window.

\subsection{UCN-Based Weight Selection}

\begin{figure}[!t]
\centering
\includegraphics[width=\columnwidth]{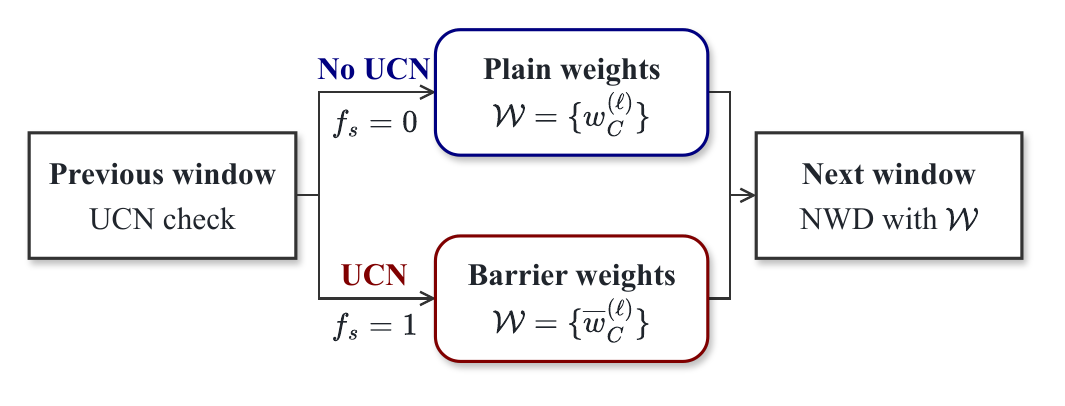}
\vspace{-6mm}
\caption{Adaptive weight switching for the next window. The UCN indicator from the previous window chooses the plain weights for $f_s=0$ and barrier weights for $f_s=1$. The first window uses the plain weights.}
\label{Fig:weight_selector}
\end{figure}

Before decoding stage $s>1$, we evaluate the syndromes of the $Mz$ lifted checks at the newly closed CN position $s-1$ using the committed hard decisions. The UCN indicator sets the error flag $f_s=1$ if any tested check is unsatisfied and $f_s=0$ otherwise. The decoder uses the barrier weights for $f_s=1$ and the plain weights for $f_s=0$, as illustrated in Fig.~\ref{Fig:weight_selector}. The same rule applies through stage $L$, except for the first stage, which uses the plain weights.

\begin{figure*}[!t]
\centering
\makebox[\textwidth][l]{%
  \hspace*{\dimexpr0.03\textwidth-3mm\relax}%
  \includegraphics[width=0.94\textwidth]{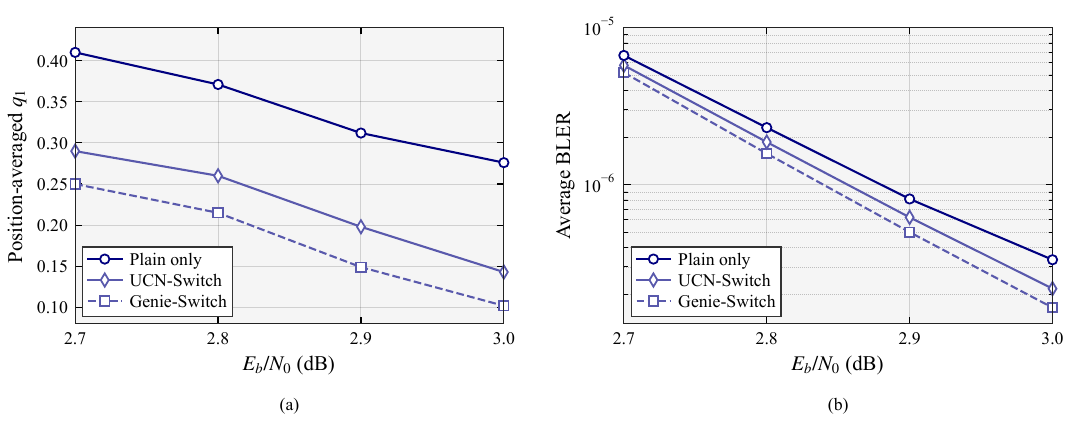}%
}
\caption{Adaptive weight switching for the $L=100$, $z=100$, $W=10$, $T=1$, $\overline{\ell}=10$ code: (a) position-averaged $q_1$ and (b) average BLER for Plain only, UCN-Switch, and Genie-Switch.}
\label{Fig:adaptive_switching}
\end{figure*}

The UCN indicator detects parity-check violations and can miss a locally valid but incorrect decision. We therefore compare NWD using only plain weights (Plain only), UCN-based weight selection (UCN-Switch), and an ideal selector using the true previous-block error event $E_{s-1}$ (Genie-Switch). Both weight sets remain fixed during decoding; the selector changes only which set is used.
For each SNR point in Fig.~\ref{Fig:adaptive_switching}, we collect frames until 500 frame errors are observed and compute average BLER and $q_1$ from the same frames. We estimate $q_1$ from true error events at adjacent block positions within the same frame and average the position-wise estimates equally over positions with observed previous-block errors. The first stage is excluded because it has no predecessor.

The position-averaged conditional error probabilities in Fig.~\ref{Fig:adaptive_switching}(a) show that adaptive weight switching suppresses failures following a previous-block error. At 3.0 dB, $q_1$ decreases from 0.276 for Plain only to 0.143 for UCN-Switch and 0.102 for Genie-Switch. The remaining gap between the two switching curves is consistent with errors that the UCN indicator does not identify.
Fig.~\ref{Fig:adaptive_switching}(b) shows the corresponding high-SNR BLER benefit. At 3.0 dB, UCN-Switch reduces the BLER from $3.33\times10^{-7}$ to $2.18\times10^{-7}$, while Genie-Switch gives $1.65\times10^{-7}$. Together, the lower conditional error probability and average BLER show that adapting the weights to the previous-block outcome improves reliability in the evaluated high-SNR range.
Algorithm~\ref{alg:NWD} summarizes online decoding with the proposed methods.

\suppressfloats[t]
\begin{algorithm}[t]
 \caption{Summary of Online NWD Decoding}
 \label{alg:NWD}
 \begin{algorithmic}[1]
 \renewcommand{\algorithmicrequire}{\textbf{Input:} }
 \REQUIRE Reduced target graph, scheduling masks for the two weight sets, plain weights $w_C^{(\ell)}$, barrier weights $\overline{w}_C^{(\ell)}$, and $\overline{\ell}$
 \FOR{window stage $s=1,\ldots,L$}
    \STATE Initialize messages using channel and inherited LLRs.
    \STATE Set $f_s=1$ if $s>1$ and any check at CN position $s-1$ is unsatisfied; otherwise, set $f_s=0$.
    \STATE Select the mask $a_C^{(\ell)}$ for the plain weights if $f_s=0$, or for the barrier weights otherwise.
    \FOR{$\ell=1,\ldots,\overline{\ell}$}
       \STATE Update VN-to-CN messages required by active CN updates ($a_C^{(\ell)}=1$) using (\ref{Eq:NWMS_VN}).
       \FOR{each retained CN update}
          \IF{$a_C^{(\ell)}=1$}
             \STATE Update CN-to-VN messages using (\ref{Eq:NWMS_CN}) with $w_C^{(\ell)}$ if $f_s=0$, else $\overline{w}_C^{(\ell)}$.
          \ELSE
             \STATE Reuse $m_{c\to v}^{(\ell-1)}$ on retained edges.
          \ENDIF
       \ENDFOR
    \ENDFOR
    \STATE Decide the target VNs, pass their output LLRs to the next window, and shift the window by one position.
 \ENDFOR
\end{algorithmic}
\end{algorithm}

\section{Complexity Analysis}\label{Sec:Complexity}

We assess computational efficiency through the message-processing operations and decoding weights required during online operation.
Table~\ref{Table:complexity} compares the operation and stored-weight counts for a nominal left-terminated reference window. Its CN degrees are $(2,4,6,\ldots,6)$, giving $E=54$ protograph edges and $E\bar{\ell}=540$ weight multiplications for the full graph. Let $d_C$ denote the degree of protograph CN $C$. The comparison count per MS-based CN update is $\alpha_C=d_C+\lceil\log_2 d_C\rceil-2$~\cite{Ryan2009}. For MS-based decoders with a schedule mask, the active-edge sum $E_a=\sum_{\ell,C}a_C^{(\ell)}d_C$ gives $A=S=2E_a$ and $P=E_a$, while comparisons sum $a_C^{(\ell)}\alpha_C$.
Real multiplications correspond to weight scaling in MS-based decoders and products of $\tanh$-transformed messages in WD\textsubscript{BP}.
For WD\textsubscript{BP}, the cost model counts $d_C$ multiplications each for prefix products, suffix products, and output combinations per CN update, giving $P=3E\bar{\ell}=1620$ for the full graph.

We use normalized costs of 1 for addition or comparison, 0.3 for sign operations, 4 for real multiplication, and 6 for a lookup-table (LUT) operation including memory access and interpolation. Normalizing the energy figures in~\cite{Horowitz2016} to a 32-bit floating-point addition gives equivalent costs of approximately 4.1 for a 32-bit floating-point multiplication and 5.6 for a 32-bit access to a 32-KB static random-access memory, motivating the multiplication and LUT coefficients of 4 and 6, respectively. The resulting costs are normalized operation-cost estimates, not measured latency or energy. The counts cover this reference window rather than a full-chain average. Training is performed offline and is excluded from the decoding cost.

\begin{table*}[!t]
\centering
\caption{Operation and stored-weight counts for a left-terminated reference window with $(WN,WM,z,\bar{\ell})=(20,10,100,10)$. Per-protograph operation counts are summed before lifting; the equivalent computational cost includes $z$.}
\label{Table:complexity}
\begin{tabular}{c|ccccc|c||c}
\hline
\multirow{3}{*}{} &
\multicolumn{5}{c|}{Complexity per protograph} &
Equivalent cost &
\multirow{3}{*}{\begin{tabular}[c]{@{}c@{}}Stored\\ weights\end{tabular}} \\
\cline{2-6}
& Addition & Comparison & Sign mul. & Real mul. & LUT
& $(A+C+0.3S$ & \\
& $A$ & $C$ & $S$ & $P$ & $\mathcal{T}$
& $+4P+6\mathcal{T})z$ & \\
\hline \hline
WD\textsubscript{BP}             & 1080 & 0   & 0    & 1620 & 1080 & 1404000 (+236\%)        & 0 \\
\hline
WD & 1080 & 610 & 1080 & 540 & 0    & 417400 (reference)      & 1 \\
\hline
NBD           & 1080 & 610 & 1080 & 540 & 0    & 417400 (reference)      & 100 \\
\hline
NWD    & 888  & 498 & 888  & 444 & 0    & 342840 (-18\%)          & 84 \\
\hline
NWD + Scheduling ($R=29$) & 568 & 316 & 568 & 284 & 0 & 219040 (-48\%) & 55 \\
\hline
NWD + UCN-Switch & 888 & 498 & 888 & 444 & 0 & 342840 (-18\%) & 168 \\
\hline
\end{tabular}
\end{table*}

\begin{figure*}[!t]
\centering
\makebox[\textwidth][l]{%
  \hspace*{\dimexpr0.03\textwidth-5mm\relax}%
  \includegraphics[width=0.94\textwidth]{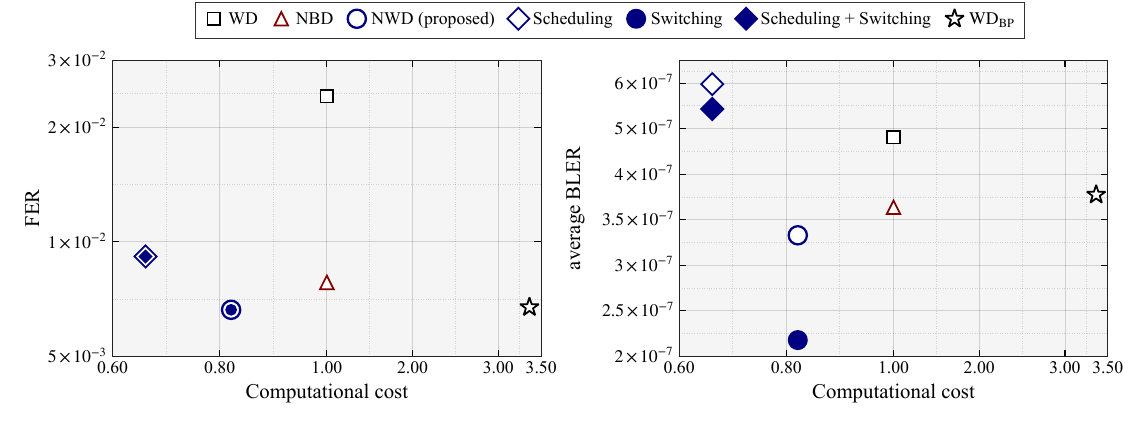}%
}
\caption{Complexity--reliability tradeoffs: left, FER at 2.4 dB; right, average BLER at 3.0 dB. Computational costs follow the model in Table~\ref{Table:complexity} and are normalized to WD. Scheduling uses $R=15$.}
\label{Fig:complexity_tradeoff}
\end{figure*}

Target-oriented decoding retains 84 of the 100 weights and reduces the estimated computational cost from 417,400 to 342,840 relative to full-graph WD. The same graph reduction also applies to WD\textsubscript{BP}, WD, and NBD; the two MS-based baselines then have the same operation counts as NWD.
With the regularized damping schedule at $R=29$, the equivalent computational cost decreases to 219,040 and 55 active weights remain. This is a further reduction of 36.1\% from NWD, or approximately 48\% relative to full-graph WD, while maintaining comparable performance. The learned damping factors are not stored during decoding, and separate training is not required for each $R$. Adaptive weight switching uses the same message updates and stores two sets of 84 weights, for a total of 168 weights, together with one selector state bit.

Fig.~\ref{Fig:complexity_tradeoff} summarizes the complexity--reliability tradeoffs using FER and average BLER. The horizontal axis gives the equivalent computational cost normalized to WD, using the reference-window cost model in Table~\ref{Table:complexity}. The scheduling results use $R=15$, providing an intermediate operating point between unscheduled NWD and the $R=29$ configuration in Table~\ref{Table:complexity}. When scheduling and switching are combined, the mask corresponding to the selected weight set is used, with 15 updates rested in each set.

Target-oriented NWD reduces the estimated computational cost by 17.9\% relative to conventional WMS window decoding while reducing FER by 72.5\%, from $2.41\times10^{-2}$ to $6.63\times10^{-3}$. Its FER is close to the $6.74\times10^{-3}$ achieved by WD\textsubscript{BP}, at a computational cost 75.6\% lower. Thus, target-oriented learning provides the error-rate improvement, while target-graph reduction removes computations that do not contribute to the required outputs. Together, they allow MS-based NWD to achieve performance comparable to sum-product window decoding at substantially lower cost.

Learned scheduling provides a further reduction in computation. With $R=15$, the cost decreases from 342,840 to 276,180, increasing the reduction relative to WD from 17.9\% to 33.8\%. The corresponding FER increases by 38.1\% relative to unscheduled NWD, to $9.16\times10^{-3}$, but remains 62.1\% lower than that of WD. Relative to WD\textsubscript{BP}, this operating point uses approximately 80.3\% less computation at a 35.9\% higher FER. Scheduling therefore provides an additional operating point that exchanges some frame-level performance for lower computational cost.

Adaptive weight switching improves average BLER by compensating for NWD's limited robustness to erroneous information inherited from previous blocks. Because the decoders behave identically until the first committed block error, adaptive weight switching does not change FER.
At 3.0 dB, switching reduces the average BLER of unscheduled NWD from $3.33\times10^{-7}$ to $2.18\times10^{-7}$, an improvement of approximately 35\%, without changing its nominal message-processing cost. Relative to WD, the average BLER reduction increases from 30.8\% for plain NWD to 54.7\% with switching. Switching also benefits the scheduled decoder: at $R=15$, its average BLER decreases from $5.98\times10^{-7}$ to $5.43\times10^{-7}$, an improvement of approximately 9.1\%. The scheduled configurations retain a higher BLER than unscheduled NWD, illustrating the remaining tradeoff between computational savings and block-level reliability.

These results demonstrate the complementary roles of the proposed methods. Target-oriented learning improves frame-level decoding, learned scheduling provides flexible computational savings, and adaptive weight switching mitigates the consequences of earlier decoding failures. Together, they exploit the structure and sequential operation of window decoding to address both computational efficiency and reliability.

\section{Conclusion}\label{Sec:Conclusion}

We proposed learning methods for SC-LDPC window decoding under a fixed iteration budget. Target-oriented training and graph reduction focus learning and computation on the emitted decisions, while learned scheduling and adaptive weight switching address selective message reuse and robustness to inherited errors, respectively. With ten iterations per window, NWD achieves performance comparable to WD\textsubscript{BP} at an estimated computational cost 75.6\% lower in the reference window. On the reduced target graph, the regularized damping-based schedule at $R=29$ lowers the estimated computational cost by approximately 48\% relative to full-graph WD in the reference window, while maintaining comparable performance at FER $10^{-1}$. Separately, UCN-based switching to weights trained on windows following decoding failures reduces average BLER by 35\% at 3.0 dB without additional decoding iterations.

The structural benefit of target-oriented decoding becomes larger when fewer iterations are combined with wider windows. Future work includes extending these methods to other codes with sequential decoding.

\section*{Acknowledgment}
The authors used OpenAI Codex~\cite{OpenAICodex} to assist with language editing and manuscript organization. The authors reviewed and approved all technical claims, mathematical content, experimental data, figures, and conclusions.

\begin{IEEEbiographynophoto}{Dae-Young Yun}
(Member, IEEE) is with the Department of Electrical, Electronic and Computer Engineering, University of Ulsan, Ulsan, South Korea. His research interests include channel coding, iterative decoding, and learning-aided communication systems.
\end{IEEEbiographynophoto}

\begin{IEEEbiographynophoto}{Hee-Youl Kwak}
(Member, IEEE) is with the Department of Electrical, Electronic and Computer Engineering, University of Ulsan, Ulsan, South Korea. His research interests include error-control coding, iterative decoding, and machine learning for communication systems.
\end{IEEEbiographynophoto}

\begin{IEEEbiographynophoto}{Sang-Hyo Kim}
(Member, IEEE) is with the Department of Electrical and Computer Engineering, Sungkyunkwan University, Suwon, South Korea. His research interests include channel coding, information theory, and communication systems.
\end{IEEEbiographynophoto}

\begin{IEEEbiographynophoto}{Yongjune Kim}
(Member, IEEE) is with the Department of Electrical Engineering, Pohang University of Science and Technology, Pohang, South Korea. His research interests include coding theory, iterative decoding, and reliable communication systems.
\end{IEEEbiographynophoto}

\begin{IEEEbiographynophoto}{Jong-Seon No}
(Life Fellow, IEEE) is with the Department of Electrical and Computer Engineering and the Institute of New Media and Communications, Seoul National University, Seoul, South Korea. His research interests include coding theory, cryptography, sequence design, and wireless communications.
\end{IEEEbiographynophoto}

\vfill
\end{document}